%% file: main.tex
\documentclass{article} 
\usepackage{iclr2027_conference,times}

\input{math_commands.tex}

\usepackage{hyperref}
\usepackage{url}
\usepackage{graphicx}
\usepackage{booktabs}
\usepackage{multirow}
\usepackage{amssymb}
\usepackage{enumitem}
\usepackage{makecell}
\usepackage{caption}
\usepackage{pifont}
\usepackage[table]{xcolor}

\newcommand*{\belowrulesepcolor}[1]{%
  \noalign{%
    \kern-\belowrulesep 
    \begingroup 
      \color{#1}%
      \hrule height\belowrulesep 
    \endgroup 
    \vspace{-0.03mm}
  }%
} 
\newcommand*{\aboverulesepcolor}[1]{%
  \noalign{%
  \vspace{-0.03mm}
    \begingroup 
      \color{#1}%
      \hrule height\aboverulesep 
    \endgroup 
    \kern-\aboverulesep 
  }%
}

\newcommand{\eg}{\emph{e.g.}}
\newcommand{\ie}{\emph{i.e.}}
\newcommand{\wrt}{\emph{w.r.t.}}
\newcommand{\cmark}{\textcolor{green!60!black}{\ding{51}}}
\newcommand{\xmark}{\textcolor{red!70!black}{\ding{55}}}

\title{Revisiting Triplet Contrastive Learning for 3D Shape Understanding}

\author{%
\parbox[t]{\dimexpr\textwidth-2\tabcolsep\relax}{%
\raggedright\normalfont
\textbf{Chaoyang Zhu}\textsuperscript{1}\hfill
\textbf{Zhe Liao}\textsuperscript{1}\hfill
\textbf{Jiayi Zhang}\textsuperscript{1}\hfill
\textbf{Jie Li}\textsuperscript{2}\hfill
\textbf{Jiayi Ji}\textsuperscript{1}\hfill
\textbf{Liujuan Cao}\textsuperscript{1,\dag}\\[0.3em]
\textsuperscript{1}Xiamen University\qquad
\textsuperscript{2}Shanghai AI Lab\\[0.6em]
}%
}

\iclrfinalcopy 
\begin{document}

\maketitle
\lhead{} 
\begingroup
\renewcommand{\thefootnote}{\fnsymbol{footnote}}
\footnotetext[2]{Corresponding author}
\endgroup

\input{abstract}
\input{introduction}
\input{related}
\input{method}
\input{experiment}
\input{conclusion}

\subsection*{Acknowledgements}

This work was supported by National Science and Technology Major Project (No. 2025YFE0113500), the National Science Fund  for Distinguished Young Scholars (No. 62525605), the National Natural Science Foundation of China (No. U25B2066, No. U22B2051, and No. 62302411), and the Xiamen Municipal Science and Technology Bureau, China (No. 3502ZC-QXT2024009).

\subsection*{AI use statement}

In this work, we used generative AI tools for supporting qualitative and thematic data analysis.
We have not used generative AI tools for other tasks with required disclosure,
and the rest of the required disclosure tasks are not applicable to this work.
Additionally, we used generative AI tools for sourcing/searching for information and editing a research paper to improve readability. We have reviewed all AI-assisted work. We take responsibility for the final content of this work, including text, claims or artifacts produced with the aid of generative AI.

\bibliography{main}
\bibliographystyle{iclr2027_conference}

\appendix
\section{Appendix}

\input{appendix}

\end{document}

%% file: math_commands.tex
\usepackage{amsmath,amsfonts,bm}

\def\eqref#1{equation~\ref{#1}}

\def\1{\bm{1}}

\DeclareMathAlphabet{\mathsfit}{\encodingdefault}{\sfdefault}{m}{sl}
\SetMathAlphabet{\mathsfit}{bold}{\encodingdefault}{\sfdefault}{bx}{n}



%% file: abstract.tex
\begin{abstract}
    In multi-modal 3D pre-training, aligning 3D shapes, their 2D image counterparts, and language descriptions has made tremendous progress. However, we find that supervision from rendered images is unreliable, the semantics encoded in synthetic views are distorted due to domain gap, self-occlusion, and rendering artifacts. Instead, we learn transferable 3D representations from language only. We observe that, in real world, 3D shapes are organized in a tree-like structure, ranging from general concepts to fine-grained instances with specific attributes, while language typically denotes generic concepts and subsumes these various 3D instantiations. Hence, we propose to embed 3D representations in hyperbolic space, whose geometry is well-suited for modeling tree-like data without distortion. To capture the asymmetric hierarchy between language and 3D shapes, we further impose a partial-order relation that language entails 3D shapes. Our method, \textbf{Hyper}bolic \textbf{C}ontrastive \textbf{L}earning (HyperCL), jointly optimizes a hyperbolic geodesic-based contrastive objective and an entailment loss to learn discriminative cross-modal representations while explicitly preserving this hierarchy. We also introduce a rigorously verified benchmark for comprehensively evaluating cross-modal 3D alignment. Across two widely used benchmarks and ours, we improve the performance up to 4.6\% absolute percentage on 3D shape classification and 5.5\% on cross-modal text-to-3D retrieval under the same parameter budget. The code and evaluation benchmark will be made publicly available.
\end{abstract}

%% file: introduction.tex

\section{Introduction}
\label{sec:intro}

\textbf{Triplet contrastive learning.} Multi-modal 3D contrastive pre-training~\citep{openshape,mrd,mm-mixing,ulip,ulip2,tamm,uni3d,mixcon3d} has emerged as a predominant paradigm for learning transferable 3D representations across various downstream tasks, \eg, zero-shot point cloud classification and cross-modal shape retrieval. However, unlike web-scale image-text data used to pretrain 2D vision-language models~\citep{clip}, high-quality language annotations for 3D shapes are extremely scarce. As a result, rendered images are often used as visual proxies for generating synthetic captions via image captioning models~\citep{blip2,gpt4}. Given these tri-modal triplets, existing methods align 3D shapes bidirectionally with language descriptions and rendered images in a shared semantic space.

While this triplet contrastive learning paradigm substantially alleviates the need for manually labeled 3D annotations, it has several limitations. Rendered images exhibit a severe domain gap \wrt~natural images used to pretrain 2D vision-language models, \ie, shifted foreground color statistics and large blank background areas~\citep{maskclip,tamm}. Furthermore, manually positioned virtual cameras following a predefined trajectory cause severe self-occlusion in certain perspectives during rendering~\citep{diffurank}. The rendering pipeline itself could also introduce unexpected errors, such as color distortions and complete rendering failures. Consequently, learning from these incomplete and misleading semantics in rendered images inevitably has a negative impact on the alignment.~\citet{tamm} attempts to address the unreliable image supervision by first adapting CLIP~\citep{clip} image encoder to rendered 3D views, thereby reducing the domain gap. However, how to avoid self-occlusion that fundamentally distorts the perceived 3D shape semantics still remains an open problem. For example, even a person may mistake a living room viewed behind its exterior wall for a rectangular box. Hence we advocate learning from language supervisions only to reduce the influence of low-fidelity and unreliable semantics in vision modality.

\textbf{Hyperbolic representation learning.} In the real world, 3D shapes often convey multifaceted semantics, ranging from various categories to intricate details like color, texture, geometry, design style, and functionality. This naturally organizes into a coarse-to-fine-grained tree that spans a wide spectrum of semantic granularities. In this hierarchy, general concepts described by language are typically positioned near the root, whereas leaf nodes represent specific 3D instantiations characterized by unique attributes. Existing approaches mainly use Euclidean space to model this tree-like structure. However, Euclidean space often struggles to preserve the underlying hierarchical structure without distortion due to its limited polynomial volume growth. As the depth of the underlying tree increases, many fine-grained concepts are compressed into a small region in the space, leading to reduced discriminative capacity. This limitation is particularly prominent in 3D shape understanding, where many real-world instances share a common coarse semantic but differ in subtle attributes.

In contrast, hyperbolic space is naturally suitable for tree-like data that has exponential volume growth \wrt its radius. This geometric property makes it useful to model an exponentially branching tree without losing discrimination capacity. Hence we propose to contrast 3D shapes against languages in hyperbolic space using a novel hyperbolic contrastive objective. The hyperbolic geometry also allows us to model the asymmetric relation between language and 3D shapes, \ie, language entails 3D shapes. To encode it explicitly, we further propose an entailment loss that ensures positive 3D representations lie inside the entailment cone of its paired text, and negative 3D shapes outside the cone. Experimental results demonstrate superior performance of our HyperCL on both zero-shot 3D shape recognition and cross-modal retrieval, enabling fine-grained cross-modal discrimination while preserving the vision-language hierarchy embedded in data. Overall, our contributions are summarized as the following:

\begin{itemize}[leftmargin=1.5em, label={-}]
    \item We identify the limitations of existing triplet contrastive learning framework, and propose to learn from language supervision only.
    \item We introduce HyperCL, a novel approach leveraging hyperbolic geometry to model the tree-like structure embedded in 3D data and enforce a partial-order relation in which language entails 3D shapes, enhancing fine-grained discrimination while preserving the hierarchy.
    \item We create a high-quality, manually verified benchmark to rigorously evaluate 3D cross-modal alignment. Experiments show that our method surpasses existing state-of-the-art by a large margin on 3D shape classification, text-to-3D, and 3D-to-text retrieval tasks.
\end{itemize}

%% file: related.tex

\section{Related Work}

\subsection{3D Triplet Contrastive Learning}

Multi-modal 3D pre-training has made tremendous progress in recent years. Besides learning from natural language supervision via bi-directional contrastive loss, the point cloud encoder also contrasts with rendered images of 3D shapes, forming a triplet of (point cloud, text, image). Due to the fact that 3D captioning is underexplored, pioneering works~\citet{openshape,ulip,ulip2,cap3d,diffurank} focus on how to accurately describe 3D shapes. They first render 3D shapes from preset viewpoints using Blender, then prompt pre-trained image captioning models~\citep{gpt4,blip2} to generate language descriptions. To cope with the scarcity of 3D pre-training data, Uni3D~\citet{uni3d} initializes its point cloud encoder directly from pretrained 2D image encoders~\citep{dino,clip,eva-clip} instead of training from scratch, establishing a strong optimization prior. MM-Mixing~\citet{mm-mixing} mixes different point clouds to learn robust and discriminative representations. MixCon3D~\citet{mixcon3d} integrates the distinct yet complementary information between rendered images and point clouds to contrast with languages. MRD~\citet{mrd} performs multi-modal distillation to maintain the semantic relations of both intra- and inter-modalities in the embedding space.

However, given rendered images, CLIP fails to encode meaningful semantics from rendered images due to domain gap. Besides, how to position virtual cameras and select appropriate viewpoints to avoid self-occlusion is still an open problem. Furthermore, rendering pipeline sometimes leads to unexpected errors. TAMM~\citet{tamm} adds an extra stage that adapts CLIP image encoder via adapter~\citep{clip-adapter} to rendered images for extracting faithful semantic information. However, better alignment with rendered images during training does not necessarily translate into better generalization, possibly due to overlapped semantics with language. Even if the distorted semantics brought by domain gap can be faithfully restored, the multi-view rendering at scale is prohibitively expensive, and the fundamental self-occlusion issue inherent to any 2D projection of 3D shapes remains difficult to address. Hence we propose to learn from language only, bypassing the limitations of the supervision signal in rendered images.

\subsection{Hyperbolic Learning}

Characterized by the exponential volume expansion, hyperbolic geometry is naturally suited for modeling hierarchical data such as trees and graphs. Hyperbolic learning can be divided into two groups: 1) the first line of work builds hyperbolic neural networks~\citep{hnn,hnn++} by designing hyperbolic counterparts of euclidean layers, such as convolution and multinomial logistic regression, and these networks are optimized using specialized riemannian optimizers; 2) the second group still keeps the main chunk of model in Euclidean space, but with an added exponential mapping projection to transform euclidean features onto hyperbolic data points. Foundational works~\citet{poincare,lorentz} establish the efficacy of two representative models in hyperbolic geometry, \ie, poincar\'e ball and lorentz model, for capturing lexical hierarchies within text modality. \citet{entailmentcone} further introduces entailment cone to encode asymmetric is-a relations in hyperbolic space, which inspires our batch-wise entailment loss.

In computer vision and multi-modal learning, hyperbolic geometry has been adapted to various tasks including 2D representation learning~\citep{HVT,HCL}, image and emotion classification~\citep{labelemo,hie,clipped}, image segmentation~\citep{his}, open-vocabulary perception~\citep{hypow,hyperlearner,hypovs}, 2D vision-language models~\citep{meru,atmg,compent}, and scaling models to billions of parameters~\citep{hypmllm}. However, applying hyperbolic geometry to model 3D data is less explored. \citet{similar} is the first to utilize hyperbolic learning in 3D domain but without large-scale pre-training. Following prior works~\citet{meru,hyperlearner} that enforce the partial relation such as object entails scenes or caption entails images/objects in 2D, we enforce caption entails 3D shapes to build vision-language hierarchy and adopt the lorentz model which is numerically more stable and suitable for gradient optimization than poincar\'e ball.

%% file: method.tex
\section{Method}

We first briefly introduce hyperbolic geometry and the triplet contrastive learning paradigm in Section~\ref{subsec:method-preliminaries}. Then we analyze the current pitfall of existing triplet contrastive learning framework in Section~\ref{subsec:method-analysis}, followed by our proposed method HyperCL with hyperbolic geodesic-based contrastive loss and entailment loss in Section~\ref{subsec:method-hypercl}.

\subsection{Preliminaries}
\label{subsec:method-preliminaries}

\textbf{Hyperbolic geometry.} Unlike Euclidean and hypersphere space with zero and constant positive curvature, hyperbolic space is a type of riemannian manifold with constant negative curvature. There are five equivalent models of hyperbolic space, and we focus on lorentz model in this work since it is suitable for optimization. We refer to~\citet{hypfoundation} for a complete treatment.

In lorentz model, hyperbolic space $\mathbb{L}^{n}$ is represented as $\left\{\mathbf{h} \in \mathbb{R}^{n+1}: \langle\mathbf{h},\mathbf{h}\rangle_{\mathcal{L}} = -1/c, c>0\right\}$ on the upper half of a two-sheeted hyperboloid. The curvature $c$ measures the degree to which hyperbolic space deviates from flat Euclidean space. A hyperbolic representation $\mathbf{h}$ is $[\mathbf{h}_{\mathrm{time}};\mathbf{h}_{\mathrm{space}}]$ with time and space component $\mathbf{h}_{\mathrm{time}} \in \mathbb{R}$, $\mathbf{h}_{\mathrm{space}} \in \mathbb{R}^{n}$. $\langle \cdot,\cdot\rangle_{\mathcal{L}}$ is lorentz inner product, computed as $\langle\mathbf{u},\mathbf{v}\rangle_{\mathcal{L}}=\langle\mathbf{u}_{\mathrm{space}},\mathbf{v}_{\mathrm{space}}\rangle_{\mathcal{E}}-\mathbf{u}_{\mathrm{time}}\mathbf{v}_{\mathrm{time}}$, and $\langle \cdot,\cdot\rangle_{\mathcal{E}}$ is euclidean inner product. Given the space component, the corresponding time component is calculated as $\mathbf{h}_{\mathrm{time}}=\sqrt{1/c+\|\mathbf{h}_{\mathrm{space}}\|^{2}}$.

Analogous to straight lines in Euclidean space, the shortest path connecting two points in hyperbolic space is called geodesic. In lorentz model, geodesics are curves traced by the intersection of the hyperboloid with hyperplanes passing through the origin $\mathbf{O}$ of $\mathbb{R}^{n+1}$. The lorentz distance between two points in hyperbolic space is:
\begin{equation}
\label{eq:lorentz-distance}
    d_{\mathcal{L}}(\mathbf{u},\mathbf{v}) = \sqrt{1/c}\cdot \cosh^{-1}\!\left(-c\langle \mathbf{u},\mathbf{v}\rangle_{\mathcal{L}}\right),
\end{equation}

The tangent space $\mathcal{T}_{\mathbf{u}}\mathbb{L}^{n}=\left\{\mathbf{h} \in \mathbb{R}^{n+1}: \langle \mathbf{h},\mathbf{u}\rangle_{\mathcal{L}} = 0\right\}$ at a hyperbolic point $\mathbf{u}$ is an Euclidean space that locally approximates hyperbolic space around $\mathbf{u}$. The exponential map is used to project a point $\mathbf{e} \in \mathbb{R}^{n}$ from tangent space onto hyperbolic manifold:
\begin{equation}
\label{eq:expm}
    \mathbf{h}_{\mathrm{space}} = \operatorname{expm}_{\mathbf{O}}(\mathbf{e}) = \frac{\sinh\!\left(\sqrt{c}\|\mathbf{e}\|\right)}{\sqrt{c}\|\mathbf{e}\|}\mathbf{e}.
\end{equation}

\begin{figure}[t]
    \centering
    \includegraphics[width=\linewidth]{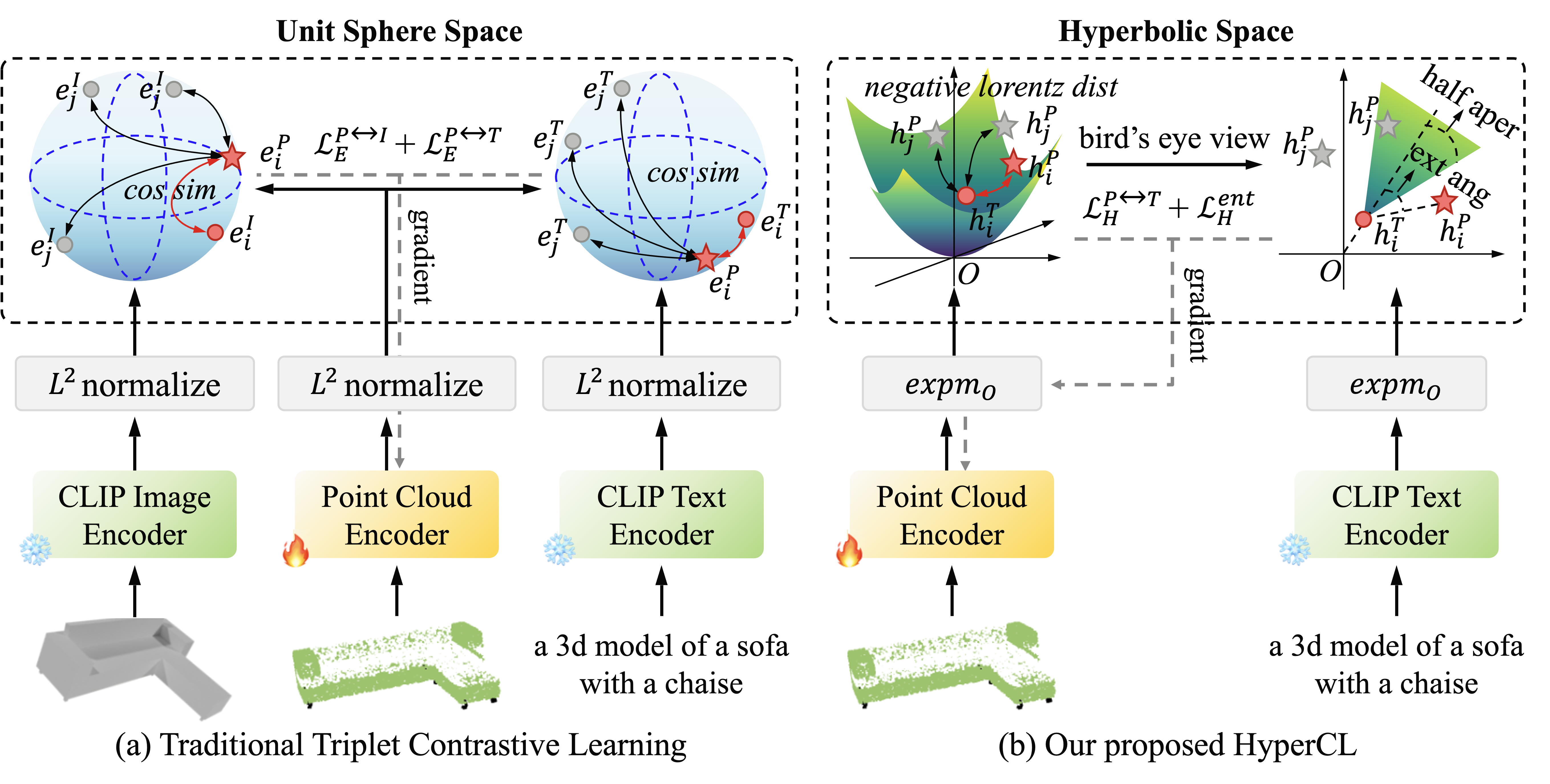}
    \caption{We embed 3D shapes in hyperbolic space to preserve exponentially expanding tree-like data. In addition, we propose an entailment loss that enforces the relation in which \emph{caption entails 3D shapes}.}
    \label{fig:arch}
\end{figure}

\textbf{Triplet contrastive learning.}
As shown in Figure~\ref{fig:arch} (a), existing methods typically take triplets $\{\mathbf{x}_{i}^{P},\mathbf{x}_{i}^{I},\mathbf{x}_{i}^{T}\}_{i=1}^{B}$ with point cloud, rendered image, and language description as input. They are first encoded into $\widetilde{\mathbf{e}}_{i}^{P}=f_{P}(\mathbf{x}_{i}^{P})$, $\widetilde{\mathbf{e}}_{i}^{I}=f_{I}(\mathbf{x}_{i}^{I})$, $\widetilde{\mathbf{e}}_{i}^{T}=f_{T}(\mathbf{x}_{i}^{T})$ by point cloud, image, and text encoders. Then they are normalized to project onto unit sphere space: $\mathbf{e}_{i}^{P}=\widetilde{\mathbf{e}}_{i}^{P}/\|\widetilde{\mathbf{e}}_{i}^{P}\|$, $\mathbf{e}_{i}^{T}=\widetilde{\mathbf{e}}_{i}^{T}/\|\widetilde{\mathbf{e}}_{i}^{T}\|$, $\mathbf{e}_{i}^{I}=\widetilde{\mathbf{e}}_{i}^{I}/\|\widetilde{\mathbf{e}}_{i}^{I}\|$. Let $l(\mathbf{a},\mathbf{b})=\log\frac{\exp(\mathcal{K}(\mathbf{a},\mathbf{b})/\tau)}{\sum_{j=1}^{B}\exp(\mathcal{K}(\mathbf{a},\mathbf{b}_{j})/\tau)}$ be the log-softmax operation where $\mathcal{K}(\cdot)$ is a similarity measure and $\tau$ is temperature. In Euclidean space, $\mathcal{K}(\cdot)$ is cosine similarity. The triplet contrastive learning objective $\mathcal{L}_{E}$ consists of the following terms:
\begin{equation}
\label{eq:euc-cts}
\mathcal{L}_{E}^{P\leftrightarrow I}
= -\frac{1}{B}\sum_{i=1}^{B}
\left(l(\mathbf{e}_{i}^{P},\mathbf{e}_{i}^{I})
+ l(\mathbf{e}_{i}^{I},\mathbf{e}_{i}^{P})\right),
\quad
\mathcal{L}_{E}^{P\leftrightarrow T}
= -\frac{1}{B}\sum_{i=1}^{B}
\left(l(\mathbf{e}_{i}^{P},\mathbf{e}_{i}^{T})
+ l(\mathbf{e}_{i}^{T},\mathbf{e}_{i}^{P})\right).
\end{equation}
Note that only the point cloud encoder is trainable, image and text encoders are directly from pretrained CLIP~\citet{clip} and they remain frozen.

\subsection{Analysis on triplet contrastive learning}
\label{subsec:method-analysis}

\begin{figure*}[t]
    \centering
    \begin{minipage}[t]{0.29\linewidth}
    \vspace{0pt}
        \centering
        \captionof{table}{We re-train OpenShape on ensembled dataset. CLIP ViT-B/16 is used as the frozen image and text encoder. Rendered images are substantially worse than languages. Acc@1 is reported.}
        \resizebox{\linewidth}{!}{%
        \begin{tabular}{cccc}
            \toprule
            Image & Text & OLVIS & MN40 \\
            \midrule
            \cmark & \cmark & 46.9 & 84.2 \\
            \cmark & \xmark & $\text{26.0}_{\color{red}{\downarrow\text{20.9}}}$ & $\text{74.2}_{\color{red}{\downarrow\text{10.0}}}$ \\
            \xmark & \cmark & $\text{45.5}_{\color{red}{\downarrow\text{1.4}}}$ & $\text{84.5}_{\color{blue}{\uparrow\text{0.3}}}$ \\
            \bottomrule
        \end{tabular}%
        }
        \label{tab:openshape}
    \end{minipage}
    \hspace{0.5pt}
    \begin{minipage}[t]{0.39\linewidth}
    \vspace{0pt}
        \centering
\includegraphics[width=\linewidth,height=3.45cm]{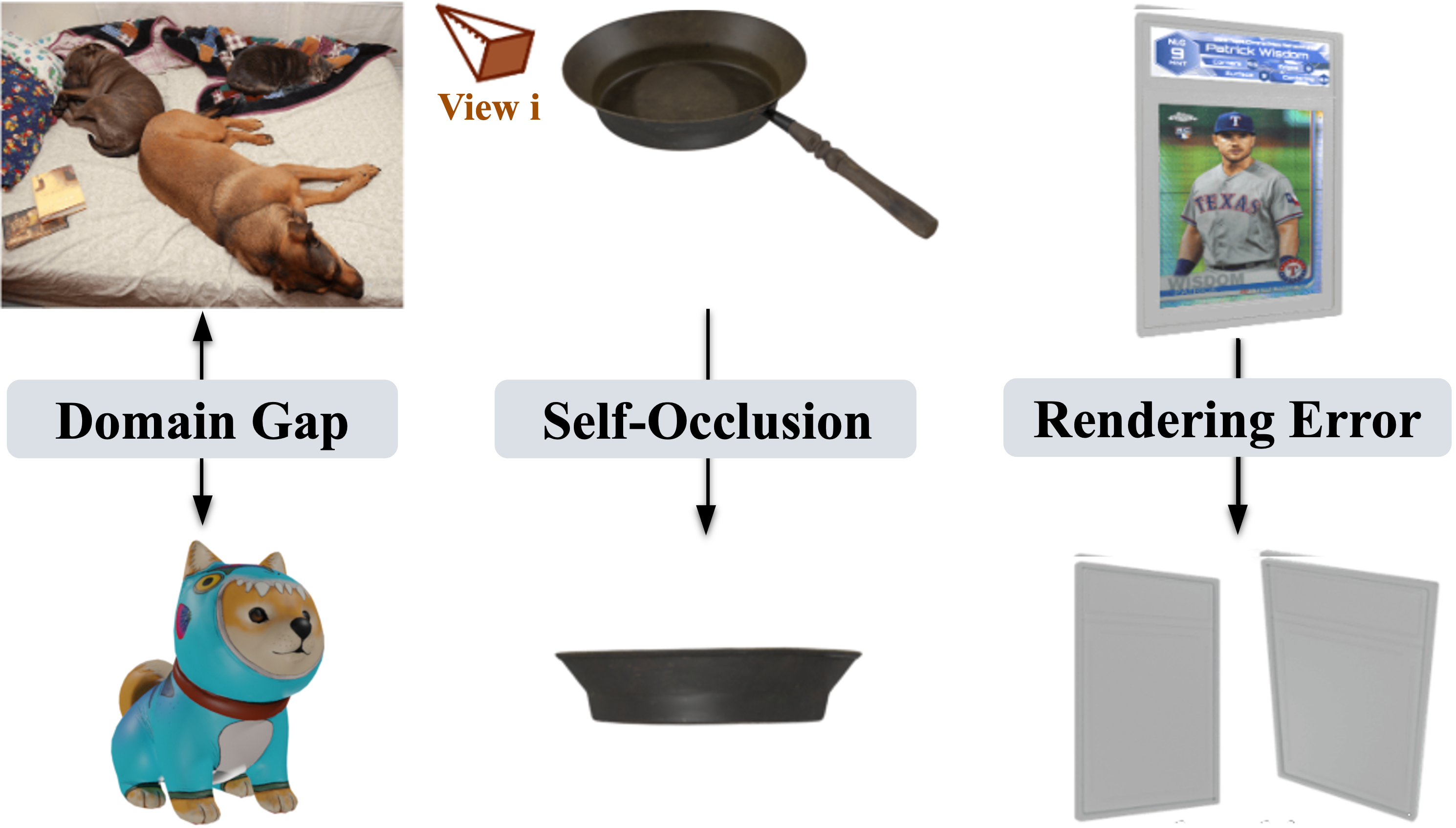}
        \captionof{figure}{Problems when contrasting with rendered images only.}
        \label{fig:problems}
    \end{minipage}
    \hspace{0.5pt}
    \begin{minipage}[t]{0.29\linewidth}
    \vspace{0pt}
        \centering
        \captionof{table}{Hyperbolicity of the representation space induced by Uni3D. A lower value indicates a stronger underlying tree-like hierarchy.}
        \resizebox{\linewidth}{!}{%
        \begin{tabular}{ccc}
            \toprule
            \multirow{2}{*}{Model} & \multicolumn{2}{c}{$\delta$-hyperbolicity $\downarrow$} \\
            \cmidrule{2-3}
            & OLVIS & MN40 \\
            \midrule
            Uni3D-B & 0.24 & 0.26 \\
            Uni3D-L & 0.20 & 0.24 \\
            Uni3D-G & 0.19 & 0.24 \\
            \bottomrule
        \end{tabular}%
        }
        \label{tab:hyperbolicity}
    \end{minipage}
    \label{fig:combined}
\end{figure*}

We investigate one representative work OpenShape~\citet{openshape}. As shown in Table~\ref{tab:openshape}, compared to triplet contrastive learning, the performance degrades severely when language supervision is removed. In contrast, after removing image supervision, it is on par with and can even marginally surpass triplet contrastive learning. 
This confirms that the gap mainly stems from the inability of CLIP to encode reliable semantics from rendered images due to domain gap~\citep{maskclip}, self-occlusion, and rendering errors. As shown in Figure~\ref{fig:problems}, the frying pan might be mistaken for a shallow bowl, and CLIP is hard to extract meaningful semantics given the rendered large blank area~\citep{maskclip}. In contrast, since language descriptions are aggregated from multiple views and inconsistent descriptions are filtered out~\citep{diffurank,ulip2}, they are more robust. Hence, we propose to learn from language supervision only.

\subsection{HyperCL}
\label{subsec:method-hypercl}

We diagnose whether the learned representation space resembles a tree structure. For this, we choose Uni3D~\citet{uni3d}, a state-of-the-art model pretrained for 3D understanding. We use $\delta$-hyperbolicity in~\citet{hie,gromov} to measure how far the space deviates from a tree, which is zero for a perfect tree and increases as the underlying structure becomes less tree-like (more details are in Appendix~\ref{subsec:appendix-hyperbolicity}). As reported in Table~\ref{tab:hyperbolicity}, the representations induced by euclidean triplet contrastive learning are already intrinsically hierarchical. This reflects the fact that real-world 3D concepts form a vast taxonomy with exponential branching factor. However, the geometry in which these representations are optimized is misaligned with this tree-like structure. Euclidean space only has a polynomial growth with insufficient discrimination capacity. In contrast, hyperbolic space can embed trees without distortion due to its exponential volume growth. Motivated by this, we embed both 3D shapes and language descriptions into hyperbolic space. We next detail our proposed HyperCL with hyperbolic contrastive loss and entailment loss.

\subsubsection{Hyperbolic Contrastive Learning.}

As shown in Figure~\ref{fig:arch} (b), we first map $\widetilde{\mathbf{e}}_{i}^{P},\widetilde{\mathbf{e}}_{i}^{T}$ onto hyperbolic space via Equation.~\ref{eq:expm}:
\begin{equation}
\mathbf{h}_{i,\mathrm{space}}^{T}=\operatorname{expm}_{\mathbf{O}}\!\left(\widetilde{\mathbf{e}}_{i}^{T}\right).
\end{equation}
Note that encoders only give space component, time component can be correspondingly derived as mentioned in Section~\ref{subsec:method-preliminaries}. In contrast to Euclidean space that uses cosine similarity, we adopt the negative lorentz distance in Equation~\ref{eq:lorentz-distance} as similarity measure $\mathcal{K}(\cdot)=-d_{\mathcal{L}}(\mathbf{u},\mathbf{v})$. Therefore, our hyperbolic contrastive loss is defined as:
\begin{equation}
    \mathcal{L}_{H}^{P\leftrightarrow T}=-\frac{1}{B}\sum_{i=1}^{B}\left(l(\mathbf{h}_{i}^{P},\mathbf{h}_{i}^{T})+l(\mathbf{h}_{i}^{T},\mathbf{h}_{i}^{P})\right).
\end{equation}
The dynamic range of negative lorentz distance is fundamentally different from the bounded $[-1,1]$ cosine similarity in Euclidean space. It is unbounded towards negative infinity, allowing for fine-grained semantics to be pushed arbitrarily far without saturation. This exponential growth in hyperbolic space naturally induces a larger discriminative capacity for the model.

\subsubsection{Batch-wise Entailment Loss.}
\label{subsubsec:entail-loss}
We further propose to enforce the relation in which language entails 3D shapes: text denotes general concepts, whereas 3D shapes are specific instantiations of those concepts. To capture this latent vision-language hierarchy, we define an entailment cone for each text $\{\mathbf{h}_{i}^{T}\}_{i=1}^B$, see Figure~\ref{fig:arch} (b) for an illustration in two dimensions. This cone is defined by half-aperture:
\begin{equation}
    \operatorname{aper}(\mathbf{h}_{i}^{T})=\sin^{-1}\!\left(\frac{2K}{\sqrt{c}\|\mathbf{h}_{i,\mathrm{space}}^{T}\|}\right),
\end{equation}
where $K$ is a small constant 0.1 used for setting boundary conditions near the origin following previous practice~\citep{meru,entailmentcone}. We now aim to identify and penalize: 1) a positive 3D shape $\mathbf{h}_{i}^{P}$ lies outside this entailment cone of its paired text; and 2) negative 3D shapes $\mathbf{h}_{j}^{P}, j=1,\ldots,B, j\neq i$ lie inside this cone. To achieve this, we measure:
\begin{equation}
    \operatorname{ext}(\mathbf{u},\mathbf{v})=\cos^{-1}\!\left(\frac{\mathbf{v}_{\mathrm{time}}+\mathbf{u}_{\mathrm{time}}c\langle\mathbf{u},\mathbf{v}\rangle_{\mathcal{L}}}{\|\mathbf{u}_{\mathrm{space}}\|\sqrt{(c\langle\mathbf{u},\mathbf{v}\rangle_{\mathcal{L}})^{2}-1}}\right).
\end{equation}
If the exterior angle $\operatorname{ext}(\mathbf{h}_{i}^{T},\mathbf{h}_{i}^{P})$ between paired text and 3D shape is smaller than $\operatorname{aper}(\mathbf{h}_{i}^{T})$, then no loss is incurred, otherwise it will be penalized to force the partial-order relation, and vice versa for unpaired 3D shapes and texts. This is captured by the following objective:
\begin{equation}
\Delta_{ij}=\operatorname{ext}(\mathbf{h}_i^T,\mathbf{h}_j^P)-\operatorname{aper}(\mathbf{h}_i^T), \qquad \mathcal{L}_{H}^{ent}=\frac{1}{B}\sum_{i=1}^{B}h(\Delta_{ii})+\frac{1}{B(B-1)}\sum_{i=1}^{B}\sum_{\substack{j=1\\j\ne i}}^{B}h(\gamma-\Delta_{ij}),
\label{eq:entailment-loss}
\end{equation}
where $h(\cdot)=\max(0,\cdot)$ is hinge function. The second term ensures that $\operatorname{ext}(\mathbf{h}_{i}^{T},\{\mathbf{h}_{j}^{P}\}_{j=1,j\neq i}^{B})\geq\operatorname{aper}(\mathbf{h}_{i}^{T})+\gamma$, thus encouraging $\{\mathbf{h}_{j}^{P}\}_{j=1,j\neq i}^{B}$ to lie outside the cone with a margin $\gamma$. Note that~\citet{meru} only imposes a pairwise entailment loss, whereas we additionally constrain negative 3D shapes in a batch to lie outside the cone.

%% file: experiment.tex
\section{Experiments}

\subsection{Implementation Details}
Following~\citet{openshape,uni3d}, we pretrain on the ensembled dataset including Objaverse~\citet{objaverse}, ShapeNet~\citet{shapenet}, 3D-FUTURE~\citet{3dfuture}, and ABO~\citet{abo}. We adopt PointBERT~\citet{pointbert} initialized from EVA02~\citet{eva02} with 85.3M trainable parameters as our point cloud encoder unless otherwise noted. EVA02-E-14-plus from EVA~\citet{eva-clip} is used as the frozen text encoder. To accelerate training, we precompute text embeddings $\{\widetilde{\mathbf{e}}^{T}\}$ and cache them offline. AdamW~\citet{adamw} with batch size 256 is adopted as the optimizer, and we train for 120 epochs. The final loss is $\mathcal{L}_{H}=\mathcal{L}_{H}^{P\leftrightarrow T}+\lambda\cdot\mathcal{L}_{H}^{ent}$, where $\lambda$ is a balancing factor and set to 0.2. The margin $\gamma$ in Equation ~\ref{eq:entailment-loss} is set to 0.5. More implementation details and training cost are in the appendix~\ref{subsec:appendix-impl} and~\ref{subsec:appendix-cost}.

\subsection{Evaluation Datasets}
We adopt two widely used datasets for assessment, including ModelNet40 (MN40)~\citet{mn40} and Objaverse-LVIS (OLVIS)~\citet{objaverse}. MN40 is a synthetic CAD model dataset. It contains 2.5K testing samples that span 40 categories. OLVIS is a subset of the Objaverse dataset with human-verified category labels. It has 46K samples spanning 1,156 categories, making it suitable for more challenging open-vocabulary and zero-shot 3D shape classification and retrieval.

\textbf{Our curated benchmark.} Astonishingly, despite its manual labeling, many samples in OLVIS are mislabeled, likely due to varied labeling quality. Such noise can lead to misleading conclusions, as it obscures whether a model truly distinguishes fine-grained semantics or simply benefits from annotation artifacts. To obtain a more reliable testbed for cross-modal alignment, we manually verify each 3D shape in OLVIS. Detailed procedures and statistics are listed in appendix~\ref{subsec:appendix-colvis-construction} and~\ref{subsec:appendix-colvis-statistics}. We term the resulting benchmark as cleaned Objaverse-LVIS (c-OLVIS), it contains around 7.7K high-quality samples across 846 categories. c-OLVIS is more trustworthy, by eliminating label noise, it ensures performance gains stem from genuine improvements in 3D semantic understanding rather than fitting to erroneous annotations, while also maintaining a large concept pool to assess real-world 3D understanding.

\subsection{Main Comparison}

\textbf{Zero-shot 3D shape classification.} We compare with strong baselines under different pre-training data regimes and model parameters in Table~\ref{tab:main-comp-cls}. We do not compare with MM-Mixing\citet{mm-mixing}, MixCon3D~\citet{mixcon3d}, and MRD~\citet{mrd} as their model weights or codes are not released. When pretrained on ShapeNet containing only 50K objects, we use a smaller model with 21.7M parameters. As shown in Table~\ref{tab:main-comp-cls}, our method achieves state-of-the-art performance across all benchmarks under the same parameter budget. Previous best model Uni3D-B achieves 81.6 on our c-OLVIS dataset, whereas we boost the performance to 86.2 with 4.6 absolute improvement. Besides, Uni3D scales its model to 306.7M (Uni3D-L) and 1B parameters (Uni3D-G) that is 3.6x and 11.9x larger than our model, however we surpass both Uni3D-L and Uni3D-G with non-trivial margin. Scaling to more parameters requires excessively long training time, and a strong regularization strategy is needed to avoid overfitting, hence we do not scale models but rather focus on improving performance via hyperbolic learning with smaller parameter budget. The strong performance indicates that our HyperCL learns more discriminative 3D representations that generalize well to challenging open-world scenarios.

\begin{table}[t]
    \centering
    \caption{Comparison on zero-shot 3D shape classification with state-of-the-art methods. To make a comprehensive comparison, we test other methods on our c-OLVIS benchmark if the model weights are released and available. Note that Uni3D-L and Uni3D-G are 3.6$\times$ and 11.9$\times$ more parameters than ours. The best is bold, and second-best is underlined.}
    \setlength{\tabcolsep}{1pt}
    \resizebox{\linewidth}{!}{%
    \begin{tabular}{lcccccccccc}
        \toprule
        \multirow{2}{*}{Method} & \multirow{2}{*}{Data} & \multicolumn{3}{c}{OLVIS} & \multicolumn{3}{c}{c-OLVIS} & \multicolumn{3}{c}{MN40} \\
        \cmidrule(lr){3-5}\cmidrule(lr){6-8}\cmidrule(lr){9-11}
        & & Acc@1 & Acc@3 & Acc@5 & Acc@1 & Acc@3 & Acc@5 & Acc@1 & Acc@3 & Acc@5 \\
        \midrule
        ReCon~\cite{recon} & \multirow{7}{*}{ShapeNet} & 1.1 & 2.7 & 3.7 & - & - & - & 61.2 & 73.9 & 78.1 \\
        CLIP2Point~\cite{clip2point} & & 2.7 & 5.8 & 7.9 & - & - & - & 49.5 & 71.3 & 81.2 \\
        ULIP~\cite{ulip} & & 6.2 & 13.6 & 17.9 & 9.9 & 19.1 & 24.1 & 60.4 & 79.0 & 84.4 \\
        OpenShape~\cite{openshape} & & 10.8 & 20.2 & 25.0 & 16.5 & 27.3 & 32.1 & 70.3 & 86.9 & 91.3 \\
        TAMM~\cite{tamm} & & \underline{13.7} & \textbf{24.2} & \underline{29.2} & \underline{23.9} & \underline{35.5} & \underline{41.2} & \underline{73.1} & \underline{88.5} & \underline{91.9} \\
        \textbf{HyperCL (ours)} & & \textbf{14.1} & \underline{24.1} & \textbf{29.2} & \textbf{25.6} & \textbf{37.2} & \textbf{43.0} & \textbf{76.0} & \textbf{89.1} & \textbf{93.4} \\
        \midrule
        \multirow{3}{*}{ULIP2~\cite{ulip2}} & \multirow{3}{*}{\makecell[c]{Objaverse\\(w/o LVIS)\\+ ShapeNet}} & & & & & & & & &  \\
        & & \underline{46.3} & - & \textbf{75.0} & - & - & - & 84.0 & - & 97.2 \\
        & & & &  & & & &  & & \\
        \midrule
        ULIP~\cite{ulip} & \multirow{6}{*}{\makecell[c]{Ensembled\\(w/o LVIS)}} & 21.4 & 38.1 & 46.0 & - & - & - & 71.4 & 84.4 & 89.2 \\
        OpenShape~\cite{openshape} & & 39.1 & 60.8 & 68.9 & 60.9 & 81.6 & 86.7 & 85.3 & 96.2 & 97.4 \\
        Uni3D-B~\cite{uni3d} & & 45.9 & \textbf{67.4} & \underline{74.8} & 72.6 & 89.2 & 92.7 & 86.1 & \textbf{97.4} & \textbf{98.7} \\
        Uni3D-L~\cite{uni3d} &  & 45.9 & 67.2 & 74.3 & \underline{73.4} & \underline{89.6} & \textbf{93.0} & \underline{86.8}  & 96.6 & 97.8 \\
        TAMM~\cite{tamm} & & 42.0 & 63.6 & 71.7 & 67.6 & 85.4 & 89.9 & 86.3 & 96.6 & 98.1 \\
        \textbf{HyperCL (ours)} &  & \textbf{46.6} & \underline{67.2} & 74.3 & \textbf{74.8} & \textbf{89.8} & \underline{92.9} & \textbf{88.3} & \underline{97.3} & \underline{98.3} \\
        \midrule
        \multirow{2}{*}{ULIP2~\cite{ulip2}} & \multirow{2}{*}{\makecell[c]{Objaverse\\+ ShapeNet}} & \multirow{2}{*}{50.6} & \multirow{2}{*}{-} & \multirow{2}{*}{79.1} & \multirow{2}{*}{78.8} & \multirow{2}{*}{92.9} & \multirow{2}{*}{95.6} & \multirow{2}{*}{84.7} & \multirow{2}{*}{-} & \multirow{2}{*}{97.1} \\
        & & & & & & & & & & \\
        \midrule
        ULIP~\cite{ulip} & \multirow{7}{*}{Ensembled} & 26.8 & 44.8 & 52.6 & - & - & - & 75.1 & 88.1 & 93.2 \\
        OpenShape~\cite{openshape} & & 46.8 & 69.1 & 77.0 & 70.3 & 90.0 & 93.5 & 84.4 & 96.5 & 98.0 \\
        Uni3D-B~\cite{uni3d} & & 51.7 & 74.0 & 80.8 & 81.6 & 95.3 & 97.7 & 86.3 & 96.5 & 97.9 \\
        Uni3D-L~\cite{uni3d} &  & 53.1 & 74.7 & 81.5 & 83.7 & 96.4 & 98.3 & 87.1 & 96.4 & 97.9 \\
        Uni3D-G~\cite{uni3d} &  & \underline{53.5} & \underline{75.3} & \underline{81.7} & \underline{84.1} & \underline{96.7} & \underline{98.3} & \underline{87.4} & \textbf{98.0} & \textbf{98.9} \\
        TAMM~\cite{tamm} & & 50.7 & 73.2 & 80.6 & 80.2 & 95.4 & 97.6 & 85.0 & 96.6 & \underline{98.1} \\
        \textbf{HyperCL (ours)} &  & \textbf{54.4} & \textbf{75.3} & \textbf{81.7} & \textbf{86.2} & \textbf{97.3} & \textbf{98.7} & \textbf{87.8} & \underline{97.1} & 98.0 \\
        \bottomrule
    \end{tabular}%
    }
    \label{tab:main-comp-cls}
\end{table}

\textbf{Zero-shot cross-modal 3D retrieval.} While prior work~\citet{uni3d} has shown the potential to perform 3D shape retrieval, we aim to quantitatively assess cross-modal 3D retrieval capabilities. To build such a dataset, we take the intersection of 3D shapes that have human written captions in Cap3D~\citet{cap3d} and those in OLVIS as test samples, leaving us with 4,421 unique shapes spanning 879 categories. As shown in Table~\ref{tab:main-comp-retr}, our model consistently achieves the best performance, with particularly clear advantages in the more challenging text-to-3D retrieval setting. Under the strict zero-shot setting where models are pre-trained on the ensembled dataset without LVIS (never seeing any test samples during training), our approach outperforms all methods, achieving a notable 3.3\% gain on text-to-3D retrieval compared to second best method TAMM~\citet{tamm}. When pretrained on full ensembled corpus, we further push the performance to 29.1. Both Uni3D and ULIP2 achieve competitive performance on 3D-to-text retrieval, but fail to maintain the same performance on text-to-3D retrieval. In contrast, due to our entailment loss, paired 3D shapes are enforced to lie inside the entailment cone, and vice versa for unpaired 3D shapes, making text-to-3D retrieval more effective while also enhancing 3D-to-text retrieval performance.

\subsection{Ablation Study}

We thoroughly investigate our proposed HyperCL in this section. To reduce training cost, the number of epochs is set to 60.

\begin{table}[t]
    \centering
    \caption{Zero-shot cross-modal 3D retrieval results. For a fair comparison, when evaluating other methods, we use their sampled point clouds from 3D meshes. Recall@K is used as evaluation metric. The best is bold, second-best is underlined. P $\rightarrow$ T denote 3D-to-text retrieval, T $\rightarrow$ P is text-to-3D retrieval. Note Uni3D-L and Uni3D-G are 3.6$\times$ and 11.9$\times$ more parameters than ours.}
    \resizebox{0.8\linewidth}{!}{%
    \begin{tabular}{lccccccc}
        \toprule
        \multirow{2}{*}{Method} & \multirow{2}{*}{Data} & \multicolumn{3}{c}{P $\rightarrow$ T} & \multicolumn{3}{c}{T $\rightarrow$ P} \\
        \cmidrule(lr){3-5}\cmidrule(lr){6-8}
        & & R@1 & R@3 & R@5 & R@1 & R@3 & R@5 \\
        \midrule
        ULIP2~\cite{ulip2} & \multirow{3}{*}{ShapeNet} & 1.0 & 2.4 & 3.6 & 1.0 & 2.7 & 3.9 \\
        TAMM~\cite{tamm} & & \underline{3.9} & \underline{9.4} & \underline{12.8} & \underline{4.9} & \underline{10.5} & \underline{14.4} \\
        \textbf{HyperCL (Ours)} & & \textbf{6.0} & \textbf{12.1} & \textbf{17.1} & \textbf{7.0} & \textbf{14.7} & \textbf{19.2} \\
        \midrule
        OpenShape~\cite{openshape} & \multirow{5}{*}{\shortstack{Ensembled\\(w/o LVIS)}} & 19.6 & 35.2 & 45.5 & 20.4 & 37.1 & 46.0 \\
        Uni3D-B~\cite{uni3d} &  & 22.0 & 39.3 & 48.8 & 18.5 & 34.1 & 42.7 \\
        Uni3D-L~\cite{uni3d} &  & 21.9 & 40.1 & 49.3 & 18.7 & 34.2 & 44.4 \\
        Uni3D-G~\cite{uni3d} &  & \underline{23.5} & \underline{40.9} & \underline{50.4} & 21.1 & 36.6 & 45.6 \\
        TAMM~\cite{tamm} & & 19.8 & 37.2 & 46.6 & \underline{21.9} & \underline{38.5} & \underline{48.0} \\
        \textbf{HyperCL (Ours)} &  & \textbf{24.6} & \textbf{42.0} & \textbf{50.9} & \textbf{25.2} & \textbf{43.4} & \textbf{52.4} \\
        \midrule
        \multirow{2}{*}{ULIP2~\cite{ulip2}} & \multirow{2}{*}{\makecell[c]{Objaverse \\+ ShapeNet}} & \multirow{2}{*}{26.3} & \multirow{2}{*}{45.2} & \multirow{2}{*}{\underline{55.1}} & \multirow{2}{*}{24.3} & \multirow{2}{*}{43.3} & \multirow{2}{*}{52.7} \\
        & & & & & & & \\
        \midrule
        OpenShape~\cite{openshape} & \multirow{6}{*}{Ensembled} & 22.2 & 39.8 & 49.3 & 23.6 & 42.5 & 52.3 \\
        Uni3D-B~\cite{uni3d} & & 24.9 & 43.2 & 53.2 & 21.7 & 39.1 & 49.0 \\
        Uni3D-L~\cite{uni3d} & & 26.0 & 44.9 & 54.8 & 22.1 & 39.4 & 48.1 \\
        Uni3D-G~\cite{uni3d} & & \underline{26.6} & \underline{45.6} & \textbf{55.5} & 22.3 & 40.5 & 50.2 \\
        TAMM~\cite{tamm} & & 23.1 & 41.2 & 51.9 & \underline{25.5} & \underline{44.1} & \underline{53.4} \\
        \textbf{HyperCL (Ours)} & & \textbf{27.6} & \textbf{45.9} & 54.8 & \textbf{29.1} & \textbf{48.4} & \textbf{57.4} \\
        \bottomrule
    \end{tabular}%
    }
    \label{tab:main-comp-retr}
\end{table}

\begin{figure*}[h]
    \centering
    \begin{minipage}[t]{0.49\linewidth}
    \vspace{0pt}\centering
    \footnotesize
    \captionof{table}{Ablation study on loss components.
    $\mathcal{L}_{E}^{P\leftrightarrow T}$ is the baseline. it does not contrast with rendered images, but adopts cosine similarity instead of negative lorentz distance. }
    \label{tab:abl-loss}
    \renewcommand{\arraystretch}{1.2}
    \begin{tabular}{lccc}
        \toprule
        & $\mathcal{L}_{E}^{P\leftrightarrow T}$ & $\mathcal{L}_{H}^{P\leftrightarrow T}$ & $\mathcal{L}_{H}^{P\leftrightarrow T} + \mathcal{L}_{H}^{ent}$ \\
        \midrule
        \belowrulesepcolor{gray!5!}
        \rowcolor{gray!5!}\multicolumn{4}{l}{\textbf{\emph{Acc@1}}}\\
        \aboverulesepcolor{gray!5!}
        \midrule
        OLVIS & 52.3 & $\text{53.2}_{\color{blue}{\uparrow\text{0.9}}}$ & $\text{53.4}_{\color{blue}{\uparrow\text{1.1}}}$ \\
        c-OLVIS & 82.9 & $\text{84.2}_{\color{blue}{\uparrow\text{1.3}}}$ & $\text{84.6}_{\color{blue}{\uparrow\text{1.7}}}$ \\
        MN40 & 85.7 & $\text{86.3}_{\color{blue}{\uparrow\text{0.6}}}$ & $\text{87.9}_{\color{blue}{\uparrow\text{2.2}}}$ \\
        \midrule
        \belowrulesepcolor{gray!5!}
        \rowcolor{gray!5!}\multicolumn{4}{l}{\textbf{\emph{R@1}}}\\
        \aboverulesepcolor{gray!5!}
        \midrule
        P$\rightarrow$T & 25.2 & $\text{26.0}_{\color{blue}{\uparrow\text{0.8}}}$ & $\text{27.0}_{\color{blue}{\uparrow\text{1.8}}}$ \\
        T$\rightarrow$P & 25.8 & $\text{27.1}_{\color{blue}{\uparrow\text{1.3}}}$ & $\text{28.4}_{\color{blue}{\uparrow\text{2.6}}}$ \\
        \midrule
        \belowrulesepcolor{gray!5!}
        \rowcolor{gray!5!}\multicolumn{4}{l}{\textbf{\emph{R@5}}}\\
        \aboverulesepcolor{gray!5!}
        \midrule
        P$\rightarrow$T & 53.4 & $\text{54.2}_{\color{blue}{\uparrow\text{0.8}}}$ & $\text{55.6}_{\color{blue}{\uparrow\text{2.2}}}$ \\
        T$\rightarrow$P & 54.7 & $\text{55.4}_{\color{blue}{\uparrow\text{0.7}}}$ & $\text{56.3}_{\color{blue}{\uparrow\text{1.6}}}$ \\
        \bottomrule
    \end{tabular}
    \end{minipage}%
    \hspace{0pt}
    \begin{minipage}[t]{0.5\linewidth}
    \vspace{0pt}\centering
    \includegraphics[width=\linewidth]{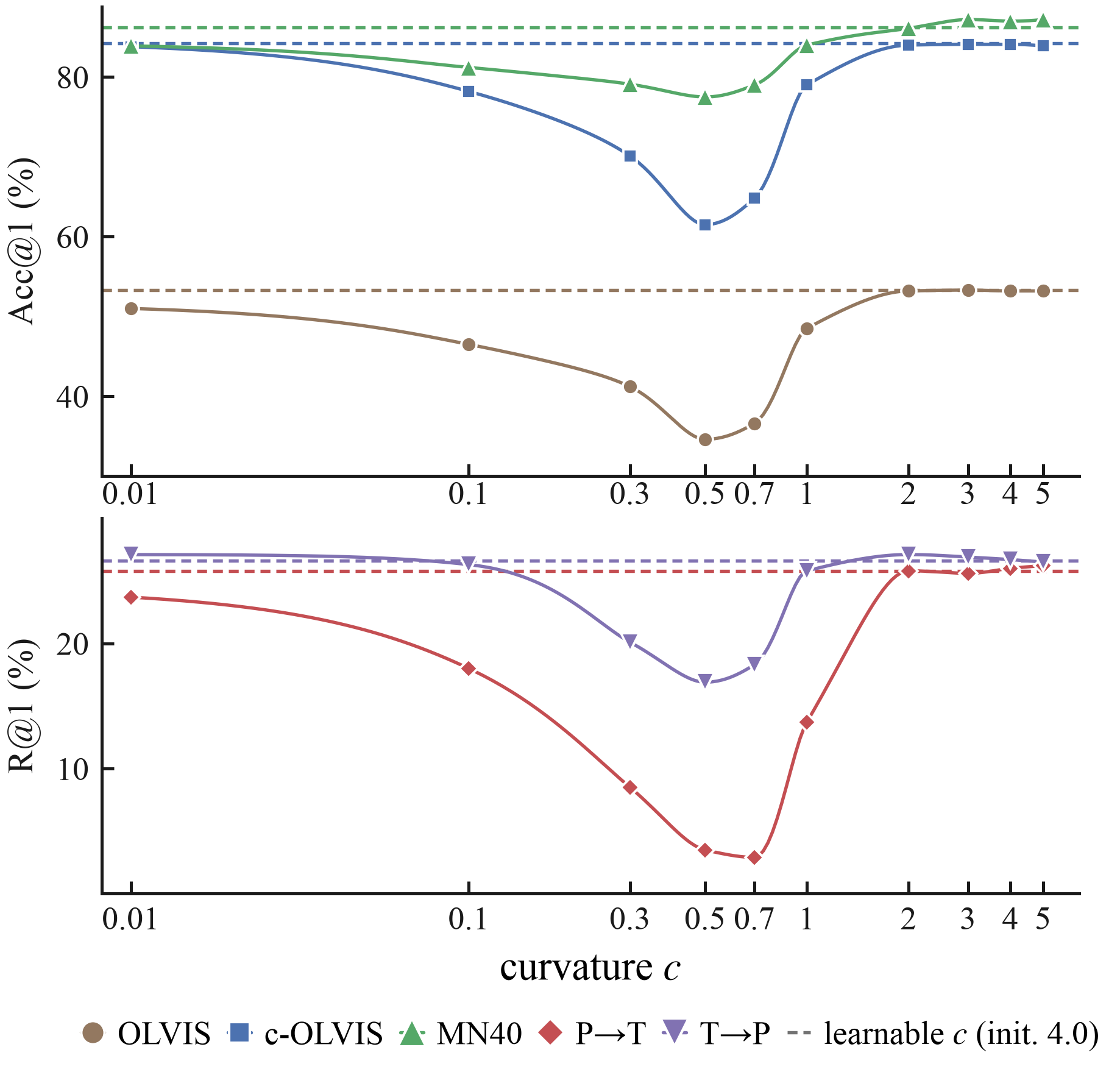}
    \captionof{figure}{Effect of curvature $c$ on loss $\mathcal{L}_{H}^{P\leftrightarrow T}$. Dashed lines denote learnable curvature.}
    \label{fig:curv}
    \end{minipage}
    \label{fig:combined-loss}
\end{figure*}

\textbf{Effect of individual loss.} Table~\ref{tab:abl-loss} shows the effect of each loss. We first compare against euclidean baseline $\mathcal{L}_{E}^{P\leftrightarrow T}$ in Equation.~\ref{eq:euc-cts}. The hyperbolic loss $\mathcal{L}_{H}^{P\leftrightarrow T}$ alone surpasses baseline by a large margin. As argued in Section~\ref{sec:intro} and~\ref{subsec:method-hypercl}, this proves that the underlying geometry is tree-like structure and is suitable to be modeled via hyperbolic geometry. The entailment loss further improves the performance. Together the proposed two losses enable discriminative recognition and partial-order relation between texts and 3D instantiations.

\textbf{Effect of curvature $c$.} The curvature measures how hyperbolic space deviates from flat Euclidean space. When it goes to zero, the hyperbolic space gradually degrades to Euclidean space. As shown in Figure~\ref{fig:curv}, both classification and retrieval performance increase from $c=1$ and stabilizes at $c\ge 2$. As $c$ decreases from 1, performance initially deteriorates, but gradually recovers when it approaches 0 as the underlying geometry becoming increasingly euclidean. However, setting $c=0.01$ still lags behind $c\ge 2$. This $v$-shaped curve suggests that moderate curvature may provide neither the strong hierarchical modeling capacity of highly curved hyperbolic space nor the near-Euclidean behavior. Making $c$ a learnable parameter (initialized from 4) is roughly the same as fixed $c=4$ since $c$ is decreasing during training and the model has to constantly adapt to this changing geometry. Hence, we set $c$ to 4 and it remains fixed.

\subsection{Qualitative Results}

\begin{figure*}[t]
    \centering
    \begin{minipage}[t]{0.58\linewidth}
    \vspace{0pt}\centering
    \includegraphics[width=\linewidth]{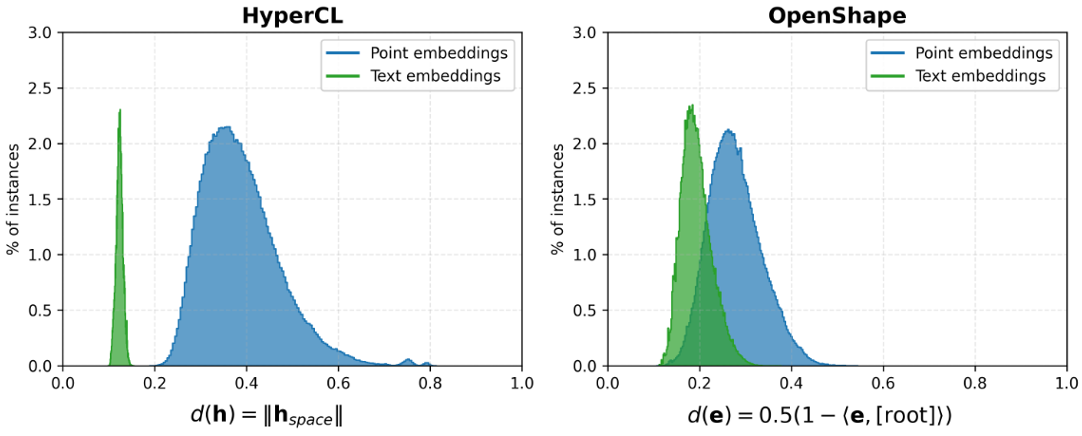}
    \captionof{figure}{Embedding distances from \texttt{[root]}. Text embeddings consistently lie closer to \texttt{[root]} than 3D shape embeddings encoded by our model.}
    \label{fig:distribution}
    \end{minipage}%
    \hfill
    \begin{minipage}[t]{0.41\linewidth}
    \vspace{0pt}\centering
    \includegraphics[width=\linewidth]{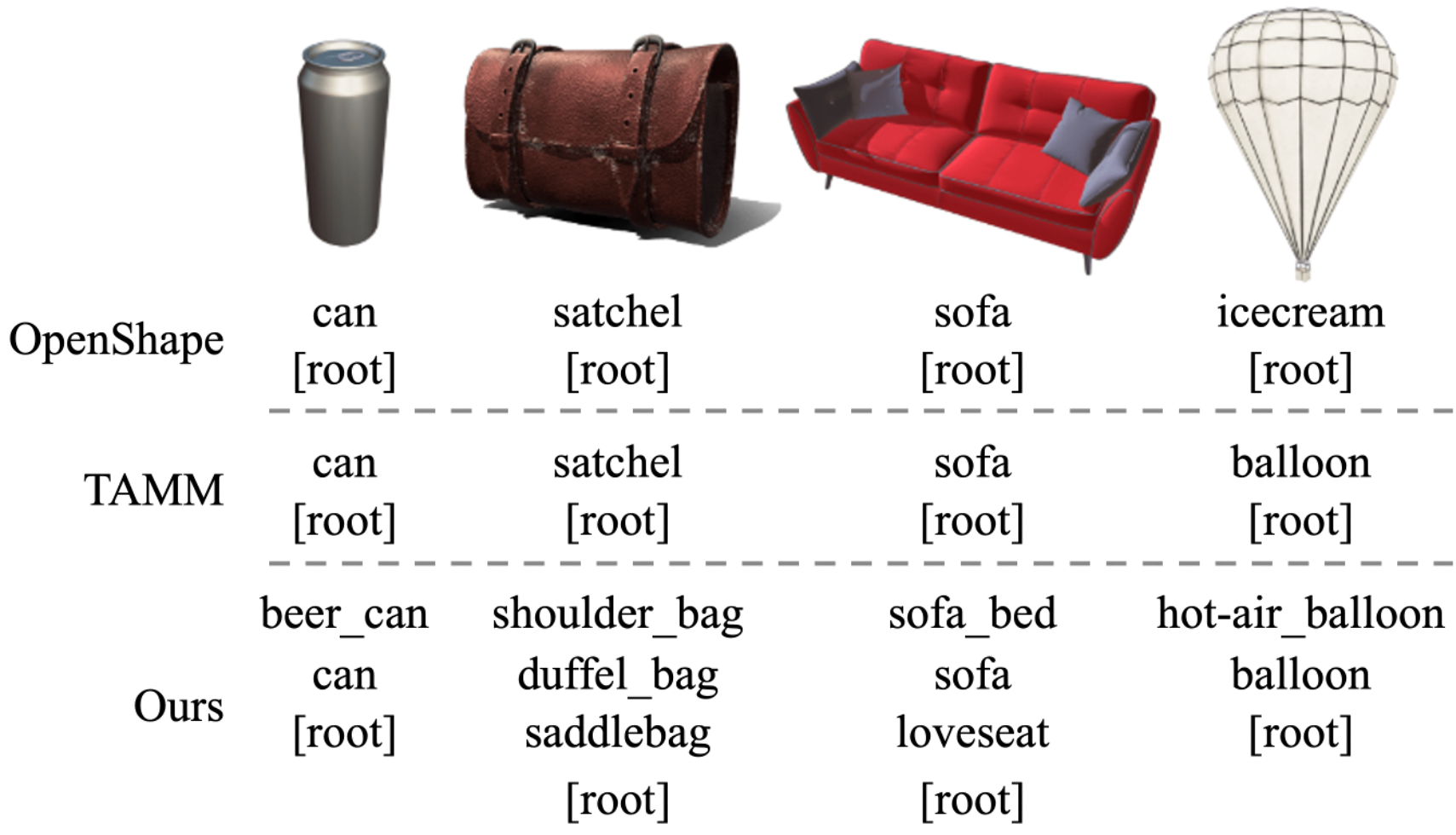}
    \captionof{figure}{Shape traversal, moving toward \texttt{[ROOT]}, our model retrieves categories at multiple levels of abstraction.}
    \label{fig:traversal}
    \end{minipage}%
\end{figure*}

If the partial-order relation of language entails 3D shapes is imposed, $h_i^T$ should lie closer to $\texttt{[ROOT]}$ than $h_i^P$, since text denotes generic concepts that subsume specific 3D instantiations. Following~\citet{meru}, the origin of lorentz model is used as $\texttt{[ROOT]}$. As shown in Figure~\ref{fig:distribution}, texts and 3D shapes are clearly separated, with text concentrated near the origin and 3D shapes scattered toward the hyperbolic boundary where the distance becomes larger to encompass fine-grained concepts. Conversely, OpenShape~\citet{openshape} inevitably overlaps two modalities with a large portion. To further examine this vision-language hierarchy, we perform shape traversal. Between a query 3D representation and origin $\texttt{[ROOT]}$, we interpolate evenly spaced points along the hyperbolic geodesic connecting them, and use each intermediate point to retrieve its nearest category. The retrieved categories should become progressively more abstract as we move toward $\texttt{[ROOT]}$. As shown in Figure~\ref{fig:traversal}, euclidean baselines~\citet{openshape,tamm} tend to directly jump to $\texttt{[ROOT]}$, whereas our model yields a much smoother trajectory of meaningful intermediate concepts, revealing multiple levels of abstraction.

%% file: conclusion.tex

\section{Conclusion}

In this paper, we present HyperCL for learning transferable 3D representations. We demonstrate that learning from language supervision only effectively bypasses the inability of VLMs to encode reliable semantics from rendered images. We leverage hyperbolic geometry to enhance discrimination and model the inherent vision-language hierarchy that language entails 3D shapes. The proposed hyperbolic contrastive and batchwise entailment losses effectively preserves this hierarchical structure. Furthermore, our manually verified c-OLVIS benchmark serves as a rigorous evaluation. Experimental results show that our method achieves state-of-the-art performance on both 3D shape classification and cross-modal retrieval tasks.

\textbf{Limitations} Though the language descriptions are relatively robust than rendered images, the problems caused by self-occlusion and rendering artifacts may still be propagated into captions. 3D captioning is best performed without rendered images as proxy via 3D LLMs that directly take 3D shapes as input. However this is beyond our paper's scope and we leave this to future work.

%% file: appendix.tex
\subsection{More Implementation Details}
\label{subsec:appendix-impl}

As listed in Table~\ref{tab:hyperparams}, we mainly follow previous practices~\citet{openshape,uni3d,ulip2} in multi-modal 3D contrastive pre-training for data augmentation and model training hyperparameters.

\begin{table}[h]
    \centering
    \caption{Hyperparameters used in our HyperCL.}
    \label{tab:hyperparams}
    \footnotesize
    \setlength{\tabcolsep}{4pt}
    \resizebox{\linewidth}{!}{%
    \begin{tabular}{r|r|r}
        \hline
        Hyperparameters & Value & Descriptions \\
        \hline
        Normalization & \checkmark & normalize xyz coordinates of point cloud into [-1, 1] \\
        Point dropout & 0.875 & the maximum ratio of dropped points is 0.875 \\
        Random scale & [0.8, 1.25] & uniformly scale points \\
        Random shift & [-0.1, 0.1] & shift x, y, z coordinates independently \\
        Random rotation & \checkmark & small rotation jittering around x, y, z axes \\
        Random rotation around gravity-axis & \checkmark & rotate around gravity-axis for learning rotation invariance \\
        Random RGB drop & \checkmark & replace the RGB values of points with 0.4 \\
        \hline
        CLIP model & ViT-E/14+ & CLIP model used to compute offline text embeddings \\
        \hline
        Number of points & 10,000 & number of sampled points from 3D mesh \\
        Hyperbolic temperature $\tau$ & 0.06 & \\
        \hline
        Learning rate & 0.001 & \\
        Learning rate schedule & WarmUpStepLR with decaying factor 0.967 & \\
        Optimizer & AdamW with betas = (0.9, 0.98) and eps=1e-6 & \\
        Weight decay & 0.01 & no weight decay on temperature $\tau$, curvature $c$ \\
        Batch Size per GPU & 256 & \\
        Epochs & 120 & \\
        AMP & \checkmark & use FP16 for training \\
        GPUs & 8 $\times$ NVIDIA A800 (PCIe, 80GB) & \\
        \hline
    \end{tabular}%
    }
\end{table}

\subsection{Training Cost}
\label{subsec:appendix-cost}

Since both our hyperbolic contrastive loss and entailment loss use the already well-optimized (inverse) hyperbolic triangle functions provided by PyTorch, and the main computation stems from point cloud encoder whose parameters still reside in Euclidean space and can be optimized via AdamW, hence both the memory footprint and FLOPs maintain the same as euclidean baselines as shown in Table~\ref{tab:train-cost}. The speed slightly decreases to 3.36 it/s compared with $\mathcal{L}_{E}^{P\leftrightarrow T}$, however, this is because the host-side kernel-launch overhead from a chain of fine-grained primitives such as \texttt{norm}, \texttt{acosh}, \texttt{clamp}, \texttt{sinh} rather than computational burden.

\begin{table}[h]
    \centering
    \caption{Comparison of training cost.}
    \label{tab:train-cost}
    \footnotesize
    \begin{tabular}{lccc}
    \toprule
        & Speed (it/s) & Peak Memory (GB) & GFLOPs \\
    \midrule
        $\mathcal{L}_{E}^{P\leftrightarrow I}+\mathcal{L}_{E}^{P\leftrightarrow T}$ & 3.30 & 46.65 & 72.05 \\
    \midrule
        $\mathcal{L}_{E}^{P\leftrightarrow T}$ & 3.43 & 46.65 & 72.05 \\
    \midrule
        $\mathcal{L}_{H}^{P\leftrightarrow T}+\mathcal{L}_{H}^{ent}$ & 3.36 & 46.65 & 72.05 \\
    \bottomrule
    \end{tabular}
\end{table}

\subsection{More Details on Hyperbolicity}
\label{subsec:appendix-hyperbolicity}

Gromov hyperbolicity quantifies how closely a metric space resembles the structure of a tree. Let $(\mathcal{M},d)$ be a metric space, where
$\mathcal{M}$ is the set of representations and
$d:\mathcal{M}\times\mathcal{M}\rightarrow\mathbb{R}$ is the distance measure.
In our experiments, each representation is a 3D shape embedding
$\tilde{e}_i\in\mathbb{R}^{D}$ produced by Uni3D~\citep{uni3d}, and $d$ is 
euclidean metric $d(\tilde{e}_i,\tilde{e}_j)=\lVert \tilde{e}_i-\tilde{e}_j\rVert_2$.

For a base point $\tilde{e}_b\in\mathcal{M}$, the Gromov product of
$\tilde{e}_i$ and $\tilde{e}_j$ is defined as
\begin{equation}
(\tilde{e}_i\mid\tilde{e}_j)_{\tilde{e}_b}
=
\frac{1}{2}
\left(
d(\tilde{e}_i,\tilde{e}_b)
+
d(\tilde{e}_j,\tilde{e}_b)
-
d(\tilde{e}_i,\tilde{e}_j)
\right).
\end{equation}
In a rooted tree, it equals the length of the initial path shared
by the paths from $\tilde{e}_b$ to $\tilde{e}_i$ and $\tilde{e}_j$.
Thus, a larger Gromov product indicates a deeper branching
point, whereas a smaller value indicates branching closer to the base point.

Given three representations $\tilde{e}_i$, $\tilde{e}_j$, and $\tilde{e}_k$, a tree satisfies the following Gromov product inequality:
\begin{equation}
(\tilde{e}_i\mid\tilde{e}_k)_{\tilde{e}_b}
\geq
\min\left\{
(\tilde{e}_i\mid\tilde{e}_j)_{\tilde{e}_b},
(\tilde{e}_j\mid\tilde{e}_k)_{\tilde{e}_b}
\right\}.
\end{equation}
The deviation from this tree condition is quantified by the base-point Gromov hyperbolicity:
\begin{equation}
\delta_{\tilde{e}_b}
=
\max_{\tilde{e}_i,\tilde{e}_j,\tilde{e}_k\in\mathcal{M}}
\left[
\min\left\{
(\tilde{e}_i\mid\tilde{e}_j)_{\tilde{e}_b},
(\tilde{e}_j\mid\tilde{e}_k)_{\tilde{e}_b}
\right\}
-
(\tilde{e}_i\mid\tilde{e}_k)_{\tilde{e}_b}
\right].
\end{equation}
A tree metric has $\delta_{\tilde{e}_b}=0$, since the above inequality holds for every triple of points. A positive value measures the largest violation of this condition among sampled representations. Therefore, a smaller value indicates a more tree-like structure, whereas a larger value indicates the opposite. To account for the overall scale of the embedding space, we report the relative hyperbolicity:
\begin{equation}
\delta_{\mathrm{rel}}
=
\frac{2\delta_{\tilde{e}_b}}
{\operatorname{diam}(\mathcal{M})},
\qquad
\operatorname{diam}(\mathcal{M})
=
\max_{\tilde{e}_i,\tilde{e}_j\in\mathcal{M}}
d(\tilde{e}_i,\tilde{e}_j).
\end{equation}
In practice, we follow~\citet{hie,gromov} to compute the hyperbolicity.

\subsection{More Details on c-OLVIS Construction}
\label{subsec:appendix-colvis-construction}

\begin{figure}[t]
    \centering
    \includegraphics[width=\linewidth]{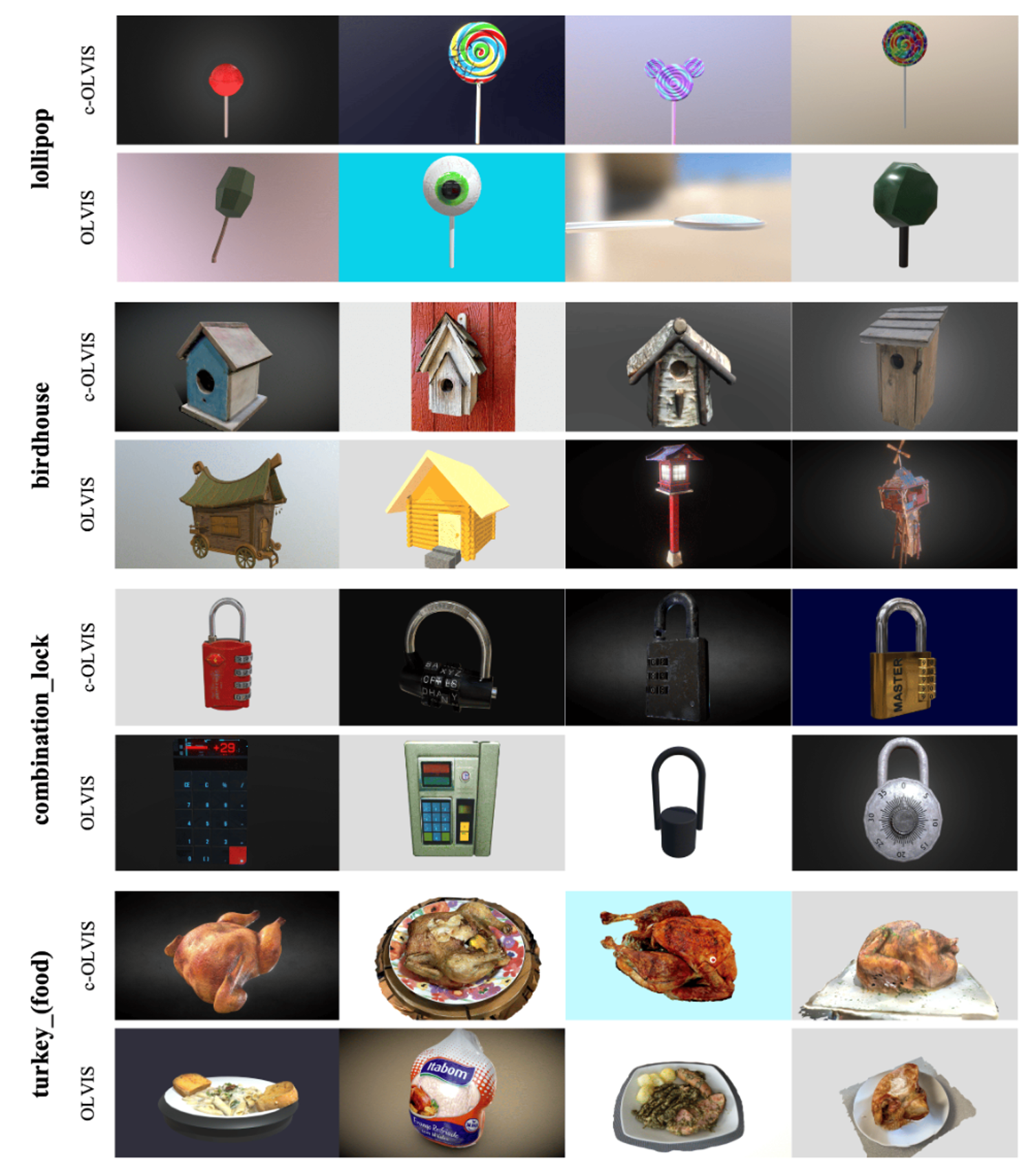}
    \caption{Examples from the manual verification. For four categories lollipop, birdhouse, combination\_lock and turkey\_(food), the upper row shows 3D shapes that were correctly labelled and kept in our c-OLVIS, while the lower row shows OLVIS samples that were mislabelled and removed.}
    \label{fig:colvis-vis}
\end{figure}

Based on OLVIS, we construct a cleaned version of it through manual inspection and refinement. During inspection, for each object we utilize the official Sketchfab website stored in the metadata. The viewer supports rotation and zooming in of the 3D shape, allowing labelers to inspect it from multiple viewpoints and reduce errors caused by self-occlusion. To ensure consistent criteria, each 3D shape will display its assigned category name, associated synonyms, along with its LVIS definition. This information helps reviewers distinguish intended category from polysemy. 

We develop a lightweight annotation tool that provides three possible options: accepted (a), rejected (r), and unsure (u). Annotators can submit decision by pressing corresponding key, and the tool automatically opens the online URL using chromium. We evenly split OLVIS into three disjoint subsets and assign each subset to one labeler. Candidates marked as accepted during initial screening phase are passed for a second round of review by other independent labelers. Only candidates accepted in both stages are kept in our c-OLVIS benchmark. This two stage procedure reduces the chances that an individual oversight or inconsistent interpretation propagates into the final benchmark. We encourage annotators to label it as unsure rather than accepted to avoid being deterministically filtered out in second round. Objects marked as unsure in the first round are collected and sent for committee screening as detailed next.

We find that many categories in OLVIS are either intrinsically ambiguous or related with each other via parent-child and synonym relationships. For example, table is the parent concept of dining table and coffee table, while in the meantime, it may also be used as a synonym for desk. We first organize such ambiguous concepts into groups via WordNet and instruct reviewers to press unsure button whenever a 3D shape lies in these groups. Then all reviewers are shown the same unsure object, when the group contains parent-children relationship, we apply a strict rule that it should be assigned to the most specific subclass unless it can not be classified into any subclass in that group, then it will be assigned as the parent class. Only the vote for a concept receives more than half of the reviewers then the object will be assigned to that class, otherwise, the object is discarded. 

We asked three graduate student to independently assess their given split using the same criteria and it takes around 15 days to annotate our c-OLVIS benchmark. The time is mostly spent on waiting for online rendering and committee review for unsure objects. Examples are given in Figure~\ref{fig:colvis-vis}.

\subsection{More Statistics of c-OLVIS}
\label{subsec:appendix-colvis-statistics}

\begin{figure*}[t]
    \centering
    \hfill
    \begin{minipage}[t]{0.27\linewidth}
    \vspace{0pt}\centering
    \includegraphics[width=\linewidth]{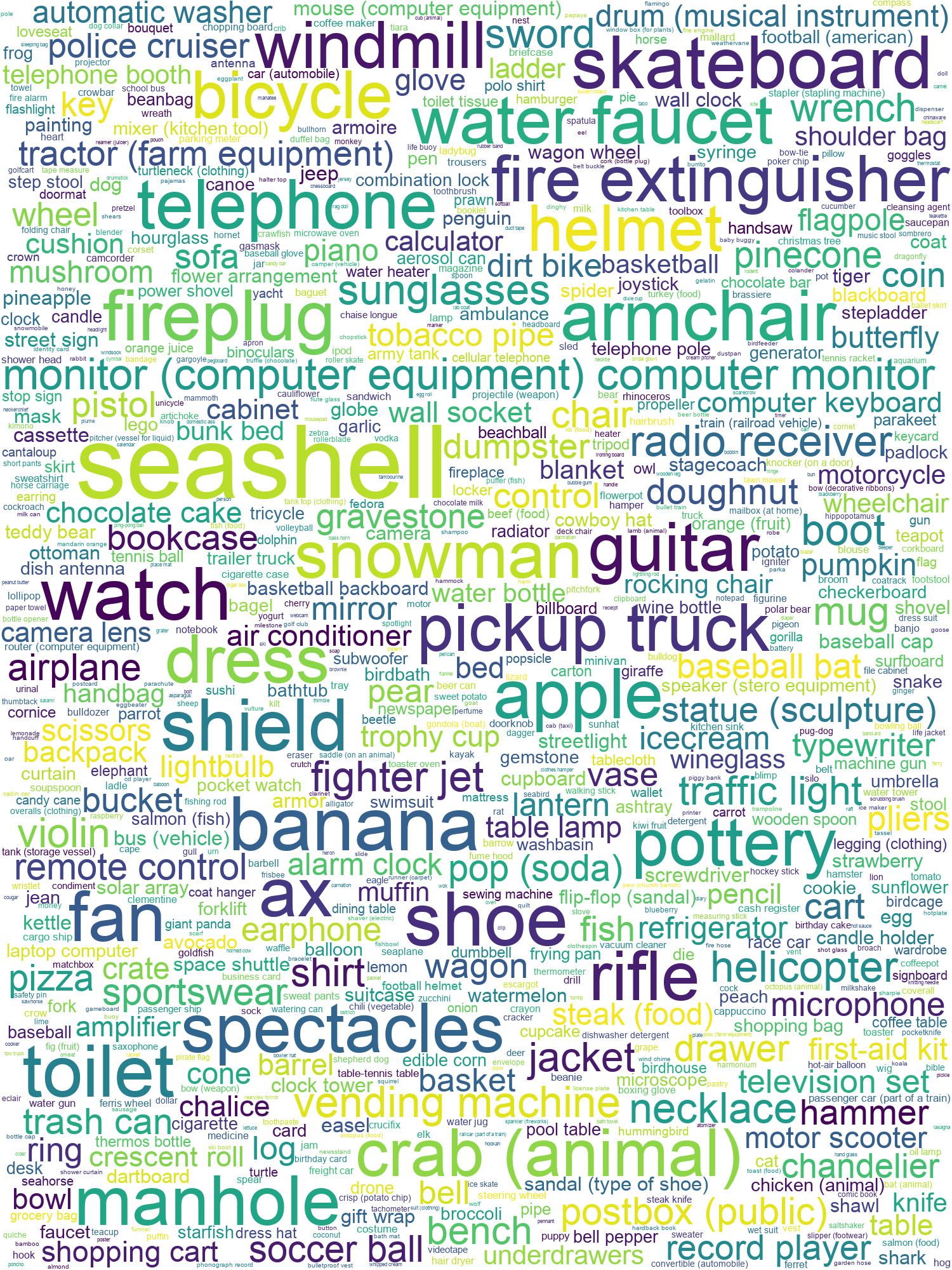}
    \end{minipage}%
    \hfill
    \begin{minipage}[t]{0.73\linewidth}
    \vspace{0pt}\centering
    \includegraphics[width=\linewidth]{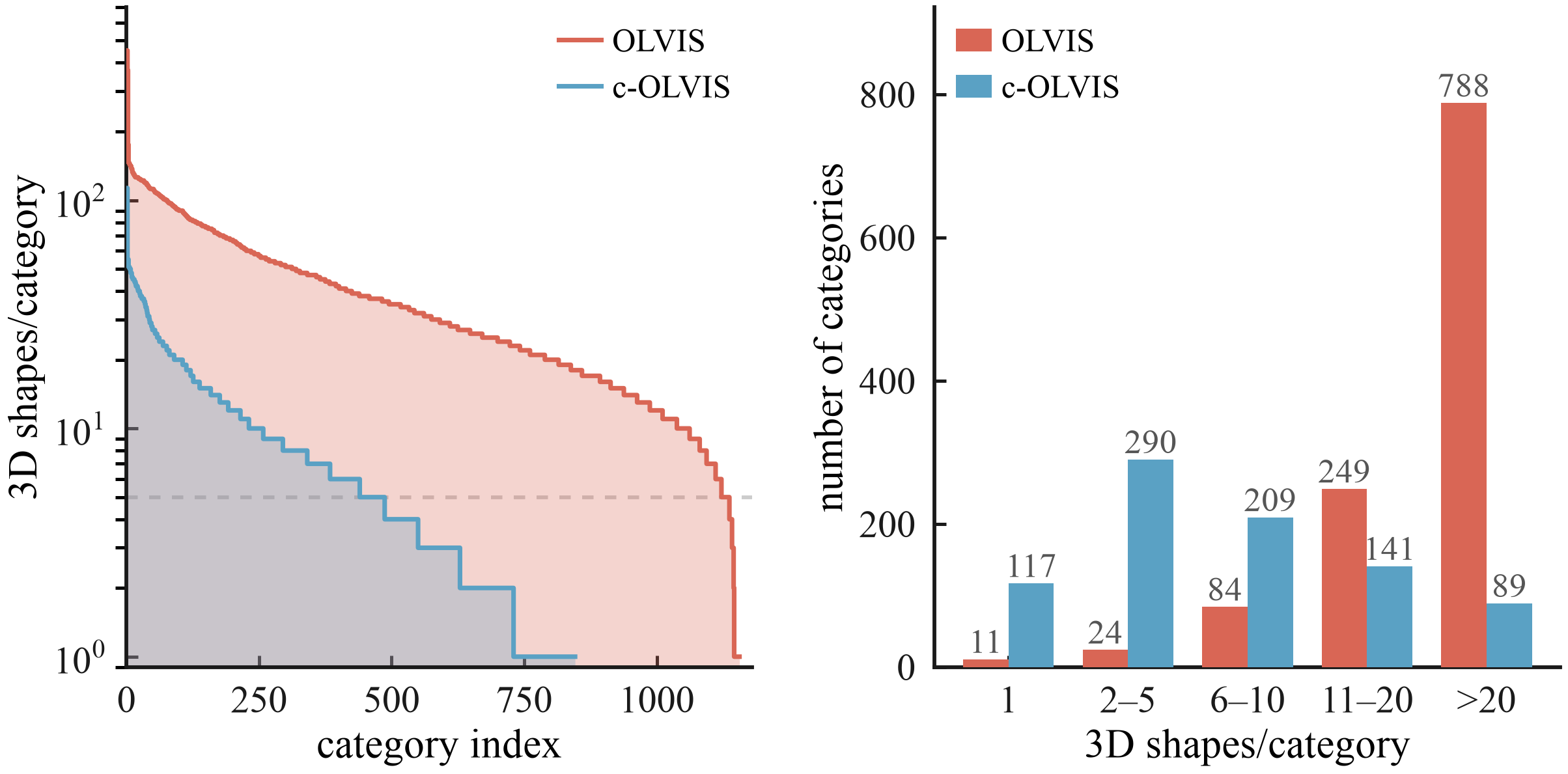}
    \end{minipage}%
    \caption{Statistics of our c-OLVIS. \textbf{Left:} category vocabulary of c-OLVIS, 846 categories in total. \textbf{Middle:} the rank--frequency distribution of class sizes on category-index axis. \textbf{Right:} the number of categories per class-size bin.}
    \label{fig:colvis-stats}
    \hfill
\end{figure*}

Compared to OLVIS, c-OLVIS has 7,678 unique 3D shapes from 846 categories, covering 73.2\% original categories of OLVIS. This diverse coverage of real-world concepts ensures that our proposed c-OLVIS benchmark remains representative for zero-shot 3D understanding. After rigorous filtering and correction, we remove substantial annotation noise, therefore it provides a more reliable evaluation, allowing the improvement to be more likely to reflect true semantic recognition rather than false exploitation on noisy labels. While at the meantime, even though the accuracy on our c-OLVIS benchmark is typically higher than that on OLVIS, the resulting c-OLVIS benchmark is not merely an easy subset that contains only frequent categories from OLVIS. As shown in Figure~\ref{fig:colvis-stats}, categories that contain less 3D shapes are well preserved, the main annotation noise comes from several categories with a large amount of objects (the rightmost bar in right subfigure). Thus, the large accuracy gap between our c-OLVIS and OLVIS can not be explained simply by retaining a large portion of easily recognizable categories. Instead, it is consistent with our manual verification process that removes many noisy or ambiguous instances while preserving low-frequency concepts. This is exactly why annotation quality matters: a reliable benchmark should reward correct semantic recognition rather than penalize it because of incorrect labels.

\subsection{More Qualitative Results}
\label{subsec:appendix-more-qualitative}

\makeatletter
\setlength{\@fptop}{0pt}
\makeatother
\begin{figure}[t!]
    \centering
    \includegraphics[width=0.75\linewidth]{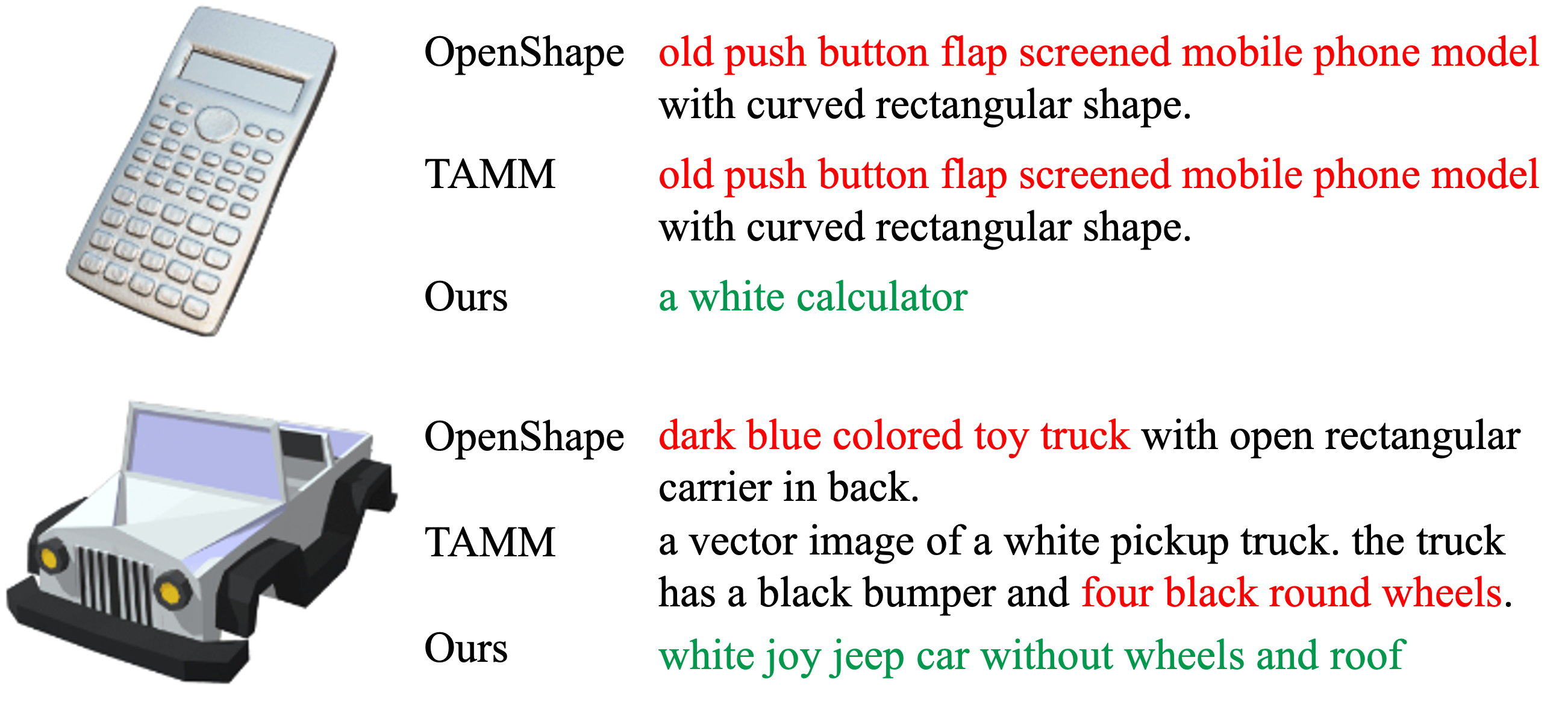}
    \includegraphics[width=0.75\linewidth]{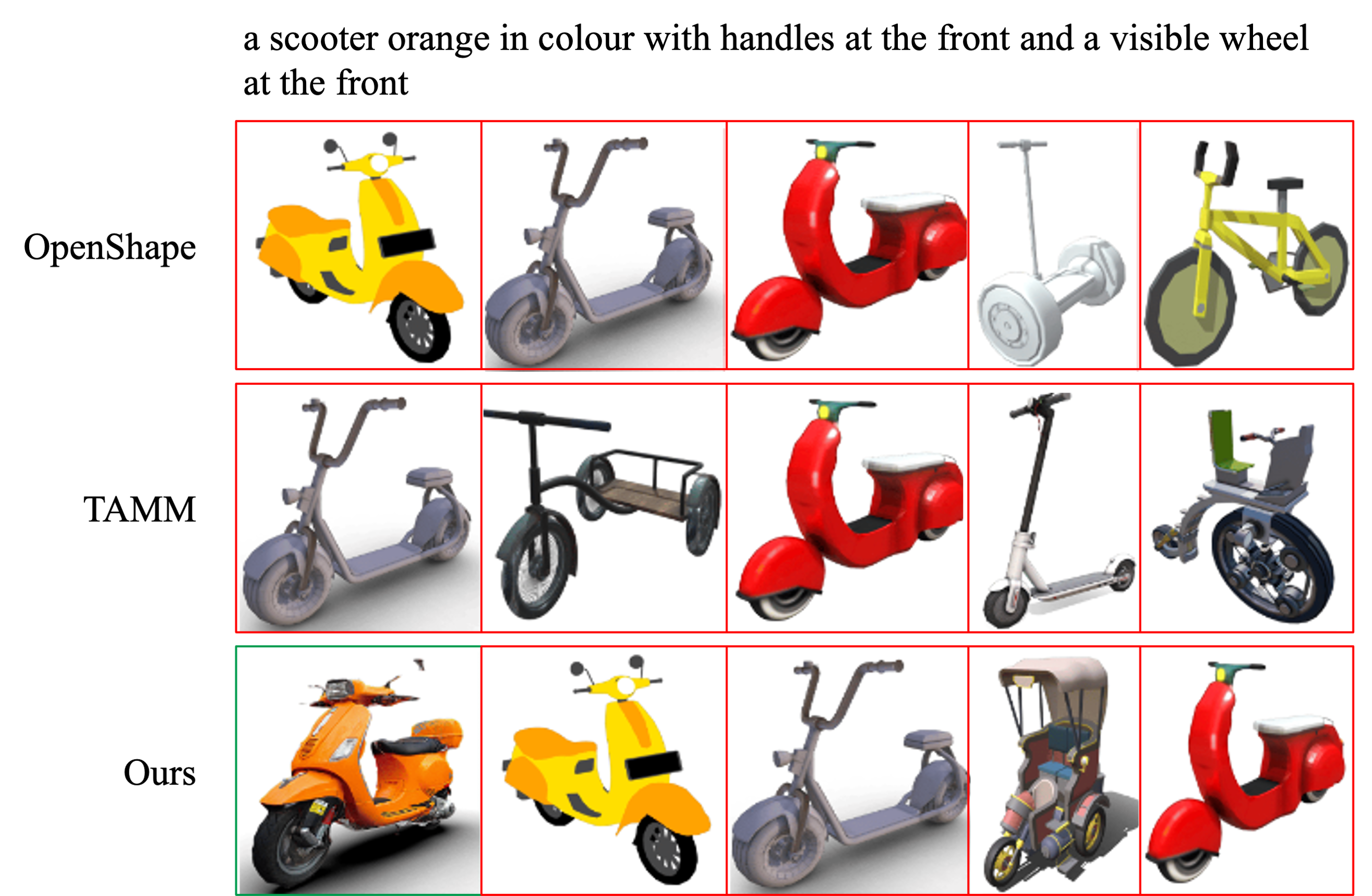}
    \caption{Qualitative visualization of retrieval results. Top: 3D $\rightarrow$ text retrieval (top-1 caption). Bottom: text $\rightarrow$ 3D retrieval (top-5 shapes). Similarities decrease from left to right.}
    \label{fig:retrieval}
\end{figure}

We visualize cross-modal retrieval results in Figure~\ref{fig:retrieval}. Euclidean geometry based models often produce partially correct retrievals, they may match coarse semantics but confuse fine-grained concepts such as car \emph{v.s.} truck \emph{v.s.} pickup truck, or miss certain attributes in the query, \emph{e.g.}, failing to recognize the orange color of the scooter. This behavior is consistent with the fact that euclidean geometry embeds tree that has a large amount of concepts with distortion, highlighting the benefit of hyperbolic space for preserving fine-grained semantics.